\documentclass[sigconf]{acmart}
\renewcommand\footnotetextcopyrightpermission[1]{} % removes footnote with conference information in first column
\usepackage{graphicx}
\usepackage{amsmath}

\usepackage{subfig}
\usepackage{booktabs}
\usepackage{multirow}
\usepackage{mathtools}
\usepackage[linesnumbered,ruled]{algorithm2e}
\usepackage{float}
\usepackage{color}
\usepackage{ulem}
\usepackage{tablefootnote}
\usepackage[capitalise]{cleveref}
\hypersetup{draft}
\AtBeginDocument{%
  \providecommand\BibTeX{{%
    \normalfont B\kern-0.5em{\scshape i\kern-0.25em b}\kern-0.8em\TeX}}}

\setcopyright{acmcopyright}
\copyrightyear{2018}
\acmYear{2018}
\acmDOI{XXXXXXX.XXXXXXX}

\acmConference[Conference acronym 'ICMR]{Make sure to enter the correct
  conference title from your rights confirmation emai}
\begin{document}

\title{TDEC: Deep Embedded Image Clustering with Transformer and Distribution Information}

\author{Ruilin Zhang}
\orcid{0000-0002-4818-9282}
\affiliation{%
  \institution{Harbin Institute of Technology, Shenzhen}
%  \streetaddress{P.O. Box 1212}
%  \city{Nanshan Qu}
  \state{Shenzhen}
  \country{China}
  \postcode{43017-6221}
}
\email{zzurlz@163.com}

\author{Haiyang Zheng}
\orcid{0000-0001-8733-9696}
\affiliation{%
	\institution{Harbin Institute of Technology, Shenzhen}
	%  \streetaddress{P.O. Box 1212}
	%  \city{Nanshan Qu}
	\state{Shenzhen}
	\country{China}
	\postcode{43017-6221}
}
\email{haiyangzheng1122@gmail.com}

\author{Hongpeng Wang}
\authornote{indicates corresponding author.}
\orcid{0000-0001-8108-2674}
\affiliation{%
	\institution{Harbin Institute of Technology, Shenzhen, \&  Peng Cheng Laboratory}
	%  \streetaddress{P.O. Box 1212}
	%  \city{Nanshan Qu}
	\state{Shenzhen}
	\country{China}
	\postcode{43017-6221}
}
\email{wanghp@hit.edu.cn}

%School of Computer Science and Technology, Harbin Institute of Technology, Shenzhen, China

%\author{Lars Th{\o}rv{\"a}ld}
%\affiliation{%
%  \institution{The Th{\o}rv{\"a}ld Group}
%  \streetaddress{1 Th{\o}rv{\"a}ld Circle}
%\email{larst@affiliation.org}
%\affiliation{%
%  \institution{Inria Paris-Rocquencourt}
%\affiliation{%
% \institution{Rajiv Gandhi University}
%\affiliation{%
%  \institution{Tsinghua University}
%  \streetaddress{30 Shuangqing Rd}
%\affiliation{%
%  \institution{Palmer Research Laboratories}
%  \streetaddress{8600 Datapoint Drive}
%  \country{USA}
%\affiliation{%
%  \institution{The Th{\o}rv{\"a}ld Group}
%  \streetaddress{1 Th{\o}rv{\"a}ld Circle}
%\email{jsmith@affiliation.org}
%\affiliation{%
%  \institution{The Kumquat Consortium}
%  \city{New York}
%\email{jpkumquat@consortium.net}

\renewcommand{\shortauthors}{Trovato and Tobin, et al.}

\begin{abstract}
% Image clustering is a crucial but challenging task in multimedia machine learning. Recently the combination of clustering with deep learning has achieved promising performance against conventional methods on high-dimensional image data. Unfortunately, existing deep clustering (DC) often ignores the importance of information fusion in a global perception scale between different image regions on clustering images, especially complex ones. Additionally, the learned features are usually clustering-unfriendly in terms of dimensionality and are based only on simple distance information for the assignment. In this regard, we propose a deep embedded image clustering TDEC, which for the first time, to our knowledge, jointly models the global dependency and reliable assignments for the image clustering task. Specifically, we introduce the Transformer to form a novel feature extraction structure T-Encoder to learn data embedding with global dependencies while using a Dim-Reduction block to achieve a clustering-friendly representation of the data embedding. Moreover, the distribution information of embedded space is considered in the clustering process to provide reliable supervised signals for joint learning. Our method is robust and allows for more flexibility in the size of the data, the number of clusters, and the complexity of the context. More importantly, the clustering performance of TDEC is higher than recent competitors. Extensive experiments with state-of-the-art approaches on benchmark and complex datasets show the superiority of TDEC.

Image clustering is a crucial but challenging task in multimedia machine learning. Recently the combination of clustering with deep learning has achieved promising performance against conventional methods on high-dimensional image data. Unfortunately, existing deep clustering methods (DC) often ignore the importance of information fusion with a global perception field among different image regions on clustering images, especially complex ones. Additionally, the learned features are usually clustering-unfriendly in terms of dimensionality and are based only on simple distance information for the clustering. In this regard, we propose a deep embedded image clustering TDEC, which for the first time to our knowledge, jointly considers feature representation, dimensional preference, and robust assignment for image clustering. Specifically, we introduce the Transformer to form a novel module T-Encoder to learn discriminative features with global dependency while using the Dim-Reduction block to build a friendly low-dimensional space favoring clustering. Moreover, the distribution information of embedded features is considered in the clustering process to provide reliable supervised signals for joint training. Our method is robust and allows for more flexibility in data size, the number of clusters, and the context complexity. More importantly, the clustering performance of TDEC is much higher than recent competitors. Extensive experiments with state-of-the-art approaches on complex datasets show the superiority of TDEC.

\end{abstract}





\maketitle

\section{Introduction}\label{Introduction}

In recent years, multimedia data, like images, have grown at a phenomenal rate and are almost ubiquitous.  Therefore, the methods or technologies to discover hidden knowledge from such data have become highly significant\cite{image2,image3}.  Image clustering aims to classify target images into clusters such that images in the same group have high homogeneity to one another and images in different groups share the maximum difference.  With the unsupervised property, image clustering plays a vital role in multimedia analysis and knowledge discovery and is widely used in image retrieval\cite{LNSCC} and image annotation\cite{Multimedia}.
%\cite{image4}

For high-dimensional, large-scale data, the conventional clustering methodologies (e.g., density-based, partitioning-based, or grid-based) would produce undesirable results due to the curse of dimensionality and shallow handcrafted features, even with the aid of PCA or LDA\cite{DDC}.
%\cite{PCA} or LDA\cite{LDA}
%\cite{DipDECK,DeepDPM,DTC}
Recently clustering paradigm combined with deep learning (called deep clustering: DC) has gained much attention because it bridges the gap between traditional methods and high dimensional data. Technically, DC methods aim to leverage unsupervised neural networks to learn the embedding representation of the raw data to help the clustering task while, in turn, using the current assignment results to further optimize the data embedding. As a pioneering work, the Deep Embedded Clustering (DEC) proposed in \cite{DEC} is the first to implement joint learning of data representation and clustering. Thanks to the simplicity of the process and the clear mathematical context, DEC and the joint training model therein have gained much attention and extension, including network backbone\cite{IDEC,LGCC,SCDCC,VaDE}, learning objectives \cite{LNSCC,ASPC-DA,DCC}, hyperparameters \cite{DipDECK,DeepDPM}, and clustering operations \cite{ICDM,DEMC}to hold multimedia data such as text \cite{DECCRL}, audio \cite{ASC-DEC}, image\cite{DEC-DA}, and numerical data\cite{DDC}.

Despite promising performance has been shown in various applications, we observe that existing methods still ignore some vital considerations when clustering image data, especially complex ones. First, existing DC methods usually do not focus on this effort of information fusion among local image regions during representation learning. Concretely, most DC methods typically employ Autoencoder(AE) or its advanced variants to enable unsupervised feature learning, in which case the learned features may be low discriminative for image clustering due to the neglect of contextual associations. While some promising works adopt convolutional networks or layers to capture the semantics of image data, the perception field in feature learning is still regional and local due to the fixed and limited kernel size, which could yield undesirable representations of complex images and thus lead to unstable clustering performance. In fact, humans perceive an object by integrating as many regional features as possible within an image to comprehend its overall semantics. For example, in feline classification, information from different image regions, such as the head, body, limbs, tail, and background, are combined to make a judgment. For this, we have a crucial insight that the information fusion with a global view between different image regions should be introduced in feature learning as such property can fully capture the discriminative information of complex images favoring the clustering task.

Second, due to the necessary regularization of unsupervised networks such as reconstruction loss, the output of embedded space (a.k.a. latent space) often remains fixed in 10 dimensions, significantly lower than the dimensionality of the raw data but still challenging for subsequent clustering behavior. In addition, most of these works adopt only simple distance information to classify the embedded features under each round of iteration. As a result, the above solutions may produce wiggling or undesirable performance for some typical scenarios such as small-scale, multi-cluster. 

With this in mind, we propose a novel deep clustering algorithm TDEC. To the best of our knowledge, this could be one of the first works to unify the Transformer, Dimension reduction, and Distribution information into the image clustering task, where the feature with global dependency, low-dimensional clustering space, and robust assignments are jointly modeled. Our work demonstrates the following contributions.

% (1) We present T-Encoder with Transformer that allows capturing discriminative information of complex images. Moreover, the proposed Dim-reduction block enables to map the feature space of the T-Encoder to a more friendly space favoring clustering behavior.

(1) We project the learning and clustering objectives in deep clustering into two different latent spaces to simultaneously achieve discriminative representation and robust partitioning. We develop the T-Encoder with Transformer that allows capturing discriminative information of complex images. Moreover, the proposed Dim-Reduction block enables to map feature space of the T-Encoder to a more friendly space favoring clustering behavior.

% To the best of our knowledge, this could be one of the first works to unify the Transformer, Dimension reduction, and Distribution information into the image clustering task, with the global dependencies, clustering-friendly representation, and reliable assignments jointly modeled. Our model could learn the discriminative low-dimensional representation of the image data and is in favor of clustering behavior, thanks to the Transformer-based encoder (T-Encoder) and the explicit dimension reduction block(DR). 

(2) We develop a Clustering Head that incorporates multi-source distribution, from which density information and neighborhood information are leveraged to improve the reliability of supervised signals as well as the final clustering performance.

(3) The proposed TDEC outperforms the newest competitors while having good robustness for challenging scenarios such as the small-scale, multi-cluster, and complex-background.

\section{Related Work}\label{Related}

The rise of deep learning has not skipped the clustering task. Initially, neural networks are often used as an alternative to traditional LDA, PCA, and LLE approaches\cite{DDC}. However, the output features are not cluster-oriented and non-representative. As the pioneering work, the deep-embedded clustering (DEC) in \cite{DEC} is then developed to perform representations and cluster assignments by end-to-end joint training simultaneously. With promising performance and flexibility, the joint learning framework proposed by DEC has been widely adopted and extended, being one of the most popular solutions among existing DC technologies.

Currently, most studies focus on the network backbone. A significant contribution in this line is given by Yin et al.\cite{IDEC}, who add reconstruction loss that acts as regularization to avoid meaningless nonlinear transformation. DEC-DA\cite{DEC-DA} introduces data augmentation, and subsequent studies \cite{SCDCC} follow this technique as well. Moreover, the work of \cite{DEC-CA} upgrades DEC by applying a convolutional AE that is beneficial for visual data, while Jiang et al.\cite{VaDE} develop a novel backbone based on a Variational AE. More complex graph convolutional AE are also considered in  \cite{LGCC}. Beyond model structure, some studies incorporate additional learning objectives in joint learning to capture more semantic information. For example, LNSCC\cite{LNSCC} deploys the positiveness and negativeness of sample pairs in a contrastive learning way to separate clusters. Likewise, the work of \cite{DECCRL} jointly learns cluster and instance-level representations by contrastive loss. Alternatively,  \cite{ASPC-DA} and \cite{IDCEC} utilize self-paced learning to dynamically update the good samples joined for training, in contrast to DCC\cite{DCC}, which introduces more labeled samples to strengthen the network based on semi-supervised.

Additionally, some work endeavors to classify the learned features in a more reasonable way, realizing the importance of pseudo-labeling (i.e., supervisory signals) under unsupervised networks. For instance, \cite{EDESC}, EDESC aims to learn the subspace bases from representation learning in an iterative manner. As the workaround to k-means, \cite{ICDM}, and \cite{DEMC} feed the learned features directly into spectral clustering, FCM fuzzy clustering, and K-median, respectively. However, these alternatives involve complex parameter settings, such as spectral clustering that requires projection size, number of neighbors, and termination threshold. To this end, some algorithms consider parameter adaptation to allow more flexibility in practice, as in the case of Leiber\cite{DipDECK}, which uses the Hartigan Dip-test to estimate the number of clusters. Instead, \cite{DeepDPM} treats the number of clusters as a network parameter and dynamically recommends them by building a specialized net. Note that most DC methods still inherit the soft assignment (maximum probability) derived from k-means, mainly because the cluster number is actually easy to estimate in practice.

% As discussed at the end of \cref{Introduction}, these methods still have some shortcomings in clustering visual data: 1) ignoring image contextual information in feature representation; 2) the dimensionality of the learned features is cluster-unfriendly and only uses common distance information in the clustering process, which may fail in contexts such as small-scale, multiple-cluster, or complex background images. Instead, our method behaves well on some complex data and outperforms current state-of-the-art methods.

As discussed at the end of \cref{Introduction}, these methods still have some shortcomings in clustering visual data: 1) ignoring image contextual information in feature representation and the dimensionality of the learned features is cluster-unfriendly;2) Using only typical distance information in the clustering process, which may fail in contexts such as small-scale, multiple-cluster, or complex background images. Instead, our method behaves well on some complex data and outperforms current state-of-the-art methods.

%Additionally, some studies have also focused on how to classify embedded features in a more reasonable way, since the pseudolabel is equally critical under unsupervised networks. For instance, in\cite{EDESC}, EDESC aims to learn the subspace bases from deep representation in an iterative refining manner. As the alternative to K-means, \cite{ICDM}, \cite{IDECF}, and \cite{DEMC} feed the embedded features directly into spectral clustering, FCM fuzzy clustering, and K-median, respectively. However, these alternatives require complex parameter settings, such as projection space size, number of neighbors, and termination threshold in spectral clustering. To this end, some algorithms take into account parameter adaptation to allow for more flexibility in practice. As in the case of Leiber\cite{DipDECK}, where use Hartigan Dip-test to estimate the number of clusters. \cite{DeepDPM}, \cite{DTC} treats the number of clusters as a network superparameter for deep embedded clustering and dynamically recommends them by building specialized sub net. Note that most deep clustering still inherits the soft assignment derived from K-means, mainly due to the simple structure and easy estimation of the number of clusters. 
%
%However, the above approaches do not focus on the role of global dependence of different regions of the image in improving feature discriminability when clustering complex visual data. Furthermore, the learned data embeddings are often unfriendly in terms of dimensionality.

\section{The Proposed Method: TDEC}\label{TDEC}
%In this section, we will delve deeper into the proposed TDEC. As shown in Figure 1, the TDEC architecture includes the TDEC Encoder, TDEC Decoder, Dimension Reduction Block, and Clustering Head. The TDEC Encoder extracts the m-dimensional picture feature, which is then reconstructed by the TDEC Decoder into an image. Simultaneously, the Dimension Reduction Block transforms the m-dimensional feature into a 2-dimensional feature. Finally, the Clustering Head is used to cluster all 2-dimensional features.
We find that feature learning and clustering processes in deep clustering have different dimensional preferences and thus project the learning and clustering objectives into two different latent spaces (termed "the feature space" and "the clustering space") to achieve discriminative representation and robust partitioning simultaneously, which has not been taken into account in the current work to the best of our knowledge.

In this paper, we aim to address the problem of grouping dataset $X \subseteq \mathbb R^d$, with $n$ samples, into $K$ disjoint clusters. The proposed TDEC architecture, as depicted in \cref{Fig1}, consists of the T-Encoder, T-Decoder, Dim-Reduction block, and Clustering Head. Given an input sample $x_i \in X$, the augmented form is denoted as $\tilde{x}_i=\mathcal{T}(x_i)$, a common augmentation function which can be any combination of random rotation, shifting, cropping, wrapping, etc. After that, the T-Encoder, denoted by $f_{\mathbf{w}}(\cdot)$, learns a $m$-dimensional feature $z_{\mathbf{w}}^i=f_{\mathbf{w}}(\tilde{x}_i)$ in feature space $\mathcal Z_w$.  The T-Decoder, represented by $g_{\mathbf{u}}(\cdot)$, then reconstructs the $m$-dimensional feature into a $d$-dimensional feature $z_{\mathbf{u}}^i$, where $z_{\mathbf{u}}^i=g_{\mathbf{u}}(f_{\mathbf{w}}(\tilde{x}_i))$. In parallel, the Dim-Reduction block transforms the $m$-dimensional feature into a 2-dimensional feature $z_{\mathbf{v}}^i=h_{\mathbf{v}}(f_{\mathbf{w}}(\tilde{x}_i))$  in the clustering space $\mathcal Z_v$. Finally, the Clustering Head incorporating the distribution information $\mathcal{Z}_v$ yields the clustering results for dataset $X$.

% Finally, the Clustering Head groups these 2-dimensional features into clusters.

\subsection{Network Architecture}\label{Networkarchitecture}

% Unlike other data types, image clustering prefers to capture visual features with local semantic information fusion, considering the content complexity and semantic coherence of images. Significantly, the self-attention mechanism enables fully connected contextual encoding across input tokens, which is strategically beneficial to enhance the clustering performance.  Based on this insight and considering the low-dimensional preference for clustering behavior, we introduce Transformer block and Dim-Reduction block to jointly learn cluster-friendly lower-dimensional representation of the data that include global semantic information. 

%To the best of our knowledge, this could be one of the first works to incorporate self- attention mechanism to deep embedded clustering architecture.
Unlike other data types, image clustering prefers to capture visual features with local semantic information fusion, considering the content complexity and semantic coherence of images. Significantly, the self-attention mechanism enables fully connected contextual encoding across input tokens, which is strategically beneficial to enhance the clustering performance.  Based on this insight, we introduce the Transformer to form a novel feature extraction structure T-Encoder to learn features with global dependency.

\begin{figure*}[h]
	\centerline{
		% \subfloat[GDD on ORL]
		{\includegraphics[width=1\linewidth]{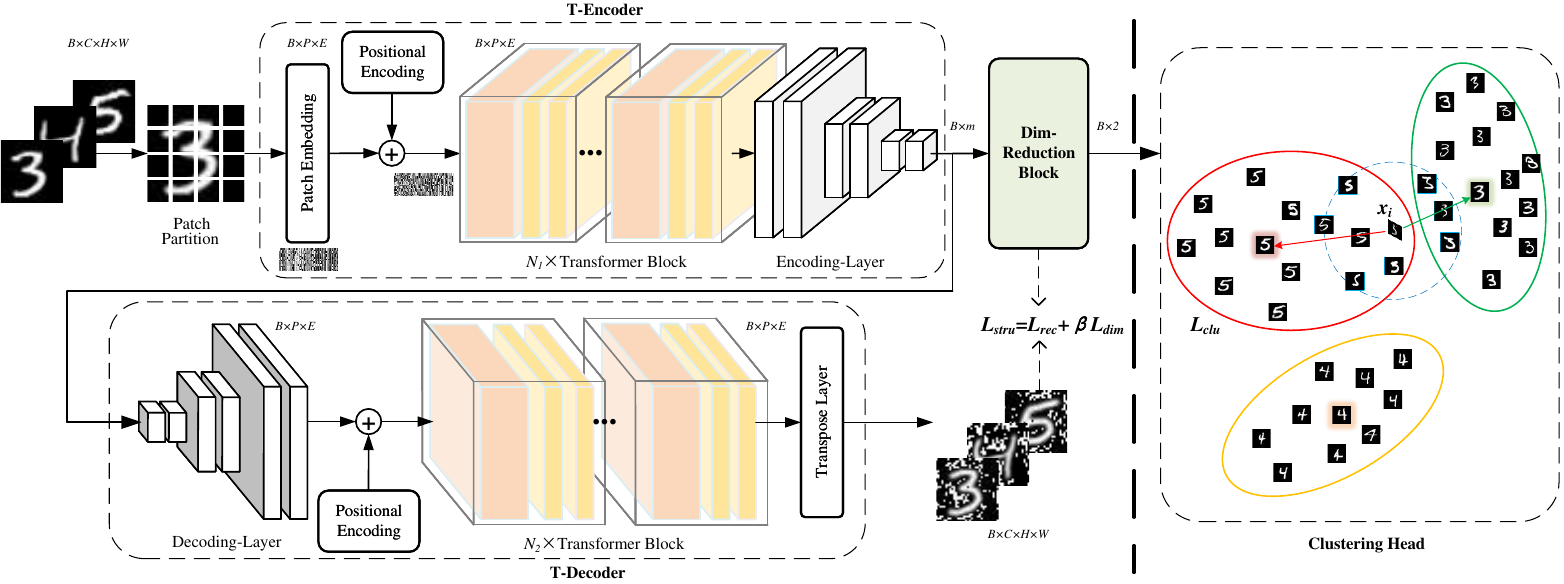}} \,}
	\caption{The framework of our proposed TDEC. It consists of T-Encoder, T-Decoder, Dim-Reduction block (DR), and Clustering Head (CH). Note that $L_{stru}$ and $L_{clu}$ represent the structure loss (the linear combination of reconstruction loss $L_{rec}$ and dimension reduction loss $L_{dim}$) and clustering loss.}
	\label{Fig1}
\end{figure*} 

\textbf{T-Encoder.} Note that the Transformer is originally oriented towards sequential data, which requires a constant latent vector size across all layers. Here we first divide the input image into non-overlapping patches to fit this principle. More specifically, we use a patch size of $\frac{H}{4}\times\frac{W}{4}$, resulting in a feature dimension of $C\times\frac{H}{4}\times\frac{W}{4}$ for the picture with input size $C\times H \times W$.  $B$, $P$, and $E$ in \cref{Fig1} denote the batch size, the number of patches, and the size of patch embeddings, respectively.

Structurally, the proposed T-Encoder consists of a Patch Embedding layer, a stack of Transformer blocks, and an Encoding-layer. The Patch Embedding Layer first converts each patch into a vector by a convolution operation. After that, the patch embeddings are combined with positional encodings (Sinusoidal-based encodings are used here) to inject information about the relative or absolute position of the input patch embeddings. 

\begin{figure}[H]
	\centerline{
		{\includegraphics[width=1\linewidth]{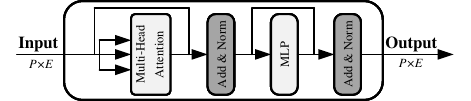}} \,}
	\caption{An individual Transformer block.}
	\label{Fig2}
\end{figure}

Instead of feeding images directly into a feature extraction backbone such as AE, TDEC first uses the linear stack of Transformer blocks to establish global dependency among different image regions to capture the global semantic information of the visual data.  Concretely, an individual Transformer block we introduced, shown in \cref{Fig2}, closely follows the design of the original Transformer Encoder. The Transformer block is built on the Multi-Head Self-Attention (MSA) module, followed by a 2-layer Multi-Layer Perceptron (MLP) with GELU nonlinearity. A Layer Normalization (LN) layer and a residual connection are added after each MSA module and MLP module. For the $l-th$ layer Transformer block, the input is $z_{\mathbf{w}}^{(l)}$, and the output is $z_{\mathbf{w}}^{(l+1)}$, computed as:
\begin{equation} \label{EQ1}
\begin{aligned}
& z_{\mathbf{w}}^{(l)\prime}=LN(MSA(z_{\mathbf{w}}^{(l)})+z_{\mathbf{w}}^{(l)})), \\
& z_{\mathbf{w}}^{(l+1)}=LN(MLP(z_{\mathbf{w}}^{(l)\prime})+z_{\mathbf{w}}^{(l)\prime}).
\end{aligned}
\end{equation}
This purposely simple stacking allows our model to be more scalable and robust, most importantly sufficient for learning embedding features in unsupervised clustering tasks. The resulting features enable focus on local patches as well as global information thanks to the self-attention mechanism. Afterwards, we further process the above features to extract the final data representation. Here, we utilize a simple but essential four linear layers, expressed as [$d$-512-512-3072-$m$], to build the feature space $\mathcal Z_w$.

%, which enhances the nonlinear characteristics of the representation.

%meet the reconstruction objective, an important constraint for unsupervised network. 

\textbf{T-Decoder.} Since there is no supervised information available to guide representation learning, adopting the principle of complete auto-encoder, the features learned by the network are reasonably constrained by the act of reconstructing them as pictures. To this aim, the T-Decoder and the T-Encoder should be approximately symmetrical in structure, consisting of a Decoding-layer, $N_2$ Transformer blocks, and a Transpose layer. Similarly, the Decoding-layer (i.e., 4 linear layers, [$m$-3072-512-512-$d$]) converts the $m$-dimension feature back to the original scale using a multi-layer network matching the Encoding-layer in T-Encoder. After the position information added, the reconstructed features also take into account global dependencies via $N_2$ Transformer blocks. Finally, feature sequences are transformed into images using deconvolution. Generally, the Transformer and related operations considered in the T-Encoder $\&$ T-Decoder enable our model to learn the discriminative features required for classifying visual data, especially with small-scale, complex contexts and multiple-cluster. The \cref{Qualitative} and \cref{Ablation} validate its effectiveness.

\textbf{Dim-Reduction Block.} To balance the difficulty of feature reconstruction and the preference of the assignment process for its dimensionality, existing DC methods typically require that the dimensionality of the embedded space should be "moderate", with most assuming 10 dimensions. Concretely, if the dimensionality of the embedded space is set too low, the reconstruction loss may fail, which in turn leads to the learned feature of the raw data irrelevant to the clustering task. Furthermore, the high-dimensional embedded space is also unfriendly as the resulting features are tough to be correctly classified in the joint learning process. 

\begin{figure}[H]
	\centerline{
		{\includegraphics[width=1\linewidth]{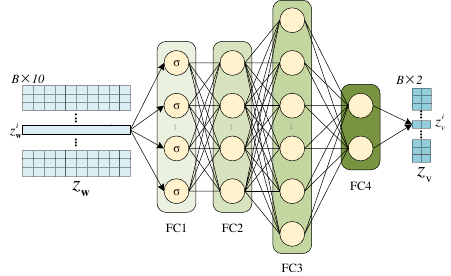}}}
	\caption{The architecture of Dim-Reduction block. Here the features $Z^i_w$ of size 1*10 learned by the T-Encoder for raw image $x_i$ is used as input to project a more cluster-friendly feature $Z^i_v$ of size 1*2.}
	\label{Fig3}
\end{figure}

Therefore, we develop a Dim-Reduction block to map the feature space $\mathcal Z_w$ generated by T-Encoder to a more friendly clustering space $\mathcal Z_v$. The Dim-Reduction block instantiates the fixed cost function in t-SNE\cite{TSNE} as a dimension reduction loss (named $L_{dim}$ in \cref{EQ11}) that can be trained and updated, which dynamically learns the distribution information under the context of dimensionality reduction along with continuously optimizing with the whole model. Since unsupervised, deep clustering models are more likely to suffer from unstable network gradients in the face of complex data, we thus do not fuse delicate modules or layers but add sophisticated linear stacks, simply but effectively, as can be verified from the ablation study\cref{Ablation}.As shown in \cref{Fig3}, with dimensions $m$-50-50-100-2.

\subsection{Clustering with Distribution Information}\label{clustering}

For the dataset $X$, TDEC utilizes the Dim-Reduction block to form a two-dimensional clustering space $\mathcal Z_v$ favoring clustering behavior. For example, for $x_i \in X$, the feature $z_v^i$ is used to represent $x_i$ in $\mathcal Z_v$, then TDEC searches for cluster centers and groups $X$ into $K$ clusters. 

Unlike recent DC methods, which directly call traditional partitioning algorithms such as the widely used k-means, we propose a Clustering Head that incorporates multi-source distribution information, from which density information and neighborhood information are leveraged to enhance clustering performance.

Specifically, density information is readily available, diverse, and promising for clustering tasks as it reflects the local structure of the data. To this end, cluster centers are instantiated as density peaks, a concept proposed in the literature\cite{DPC}, which holds that cluster centers are those points with larger densities and at a greater distance from high-density points, and formalizes them as density peak points. The density peaks prefer to be stable and better characterize a cluster than the geometric centers formed by the mean statistics used in traditional clustering like k-means. 
\begin{equation} \label{EQ2}
\begin{aligned}	
	\rho_i = \sum^{n}_{j=1}exp(-\frac{\|z_{\mathbf{v}}^{i}-z_{\mathbf{v}}^{j}\|^2_2}{{dc}^2})
\end{aligned}
\end{equation}	
where \cref{EQ2} is a common gaussian kernel density and $\rho_i$ denotes density value of $x_i$ under clustering space $\mathcal Z_v$. The $dc$ denotes the bandwidth, which can be empirically set to meet that the average number of neighbors is $2\%$ of the total number of points in $X$.

After that, the minimum distance of object $x_i$ in clustering space $\mathcal Z_v$, denoted as $\delta_i$, is defined as the smallest distance between $x_i$ and the other points whose density is higher than that of $x_i$.
\begin{equation} \label{EQ3}
\textbf{\textit{ {$ \delta_i = \left\{ \begin{gathered}
			\min \{ \left. {dist(z_v^i,z_v^j)} \right|\rho_i < \rho_j, x_j \in X\}  \;\;\\	if\;\exists x_j \in X,\;\rho_i < \rho_j \hfill \\
			%			\hfill \\
			\max \{ \left. {dist(z_v^i,z_v^j)} \right|x_j \in X\} \;\;\;{otherwise} \hfill \\ 
			\end{gathered}  \right. $}                                                                         }}.
\end{equation}
Next, the decision value $\gamma_i$ of cluster center is given by \cref{EQ4}.
\begin{equation} \label{EQ4}
\gamma_i = \rho_i*\delta_i,
\end{equation}
where $*$ is a plain product operation that acts as a magnifying discrete. In this context, given the number of clusters $K$, the true cluster centers have the top $K$ largest $\gamma$ scores and thus have good guidance for the cluster assignment.

It is well known, additionally, that a cluster act as a local bound against data space and that the objects within them are highly homogeneous. In social networks, it is common human perception that some important attribute information about a person can be derived almost unbiasedly in an objective way based on his/her close friends. In view of this, we focus on the role of neighborhood information in the partitioning process, expressed as follows.		
\begin{equation} \label{EQ5}
\theta_i^{(s)}=\frac{({\|z_{\mathbf{v}}^{s}-z_{\mathbf{v}}^{i}\|^2_2})^{-1}}{\sum_{s^{\prime} \in knn(i)}({\|z_{\mathbf{v}}^{s^{\prime}}-z_{\mathbf{v}}^{i}\|^2_2})^{-1}},	
\end{equation}	
where $\theta_i^{(s)}$ denotes the degree to which $x_i$ is influenced by its neighbor $s$ in terms of belonging clusters. The $knn(i)$ is the set of neighbors of $x_i$ determined by k-Nearest Neighbors. The purpose of computing $\theta$ is to dynamically incorporate neighbor influence in the soft assignment process, i.e., to estimate the exact distance between the target object and each cluster using its neighbors instead of itself. Formally, as given by \cref{EQ6}

\begin{equation} \label{EQ6}
q_{it} = \frac{\sum_{s \in knn(i)} \theta_i^{(s)}*(1+\|z_{\mathbf{v}}^{s}-c_t\|^2)^{-1}}{\sum_{t^{\prime}}\sum_{s \in knn(i)} \theta_i^{(s)}*(1+\|z_{\mathbf{v}}^{s}-c_{t^{\prime}}\|^2)^{-1}},
\end{equation}
where $q_{it}$ is the probability that object $x_i$ belongs to cluster $C_t$ by Student's t-distribution\cite{DECCRL}. Particularly, instead of directly minimizing the pairwise distance of object $x_i$ and center $c_t$ as before, our model jointly accounts for the role of surrounding neighbors.

The right side of \cref{Fig1} depicts the role of neighborhood information in clustering assignment, where the three clusters contain characters 5, 4, and 3, respectively, and the ones with halos are the cluster centers. Image $x_i$ (see parallelogram) is scrawled, and if we rely only on typical distance information, then $x_i$ would be incorrectly identified as "3" because of the closest distance to the cluster center of cluster 3. Alternatively, $x_i$ shares 8 neighbors ($k$=8), 5 of which belong to cluster "5" and 3 to cluster "3". By \cref{EQ5} and \cref{EQ6}, the probability that $x_i$ belongs to cluster "5" is the highest ($q_{i3}$=0.21, $q_{i4}$=0.24, $q_{i5}$=0.55), which is in line with the actual. Algorithmically, the distribution information from the density and neighborhood aspects boosts the final clustering performance of our model thanks to the clustering Head allows for more flexibility in the data structure, which can be verified from \cref{Experiments}. Note that the number of nearest neighbors $k$ is empirically set to 50 in this work, and the robustness is verified in \cref{Sensitivity}.

\subsection{Optimization}\label{optimization2}
Conceptually, there are four kinds of parameters to optimize or update: T-Encoder weights $\mathbf w$, T-Decoder weights $\mathbf u$, Dim-Reduction block weights $\mathbf v$ and target distribution $\mathbf{P}$ described below. Given an embedding $z_v^i \in \mathcal Z_v$ of $x_i$, the clustering Head that incorporates the distribution information generates its soft assignment probability $q_{it}$ for each cluster $C_t$, $t=[1,2,...,K]$.

To iteratively estimate the true assignments, we adopt the self-training strategy similar to IDEC\cite{IDEC}. This process starts by calculating a target cluster assignment $\mathbf{Q}$ from $\mathbf{P}$ by applying a certain normalization on $\mathbf{P}$, and then minimizing the clustering loss $L_{clu}$ by the Kullback-Leibler(KL) divergence between the distributions $\mathbf{P}$ and $\mathbf{Q}$:
	\begin{equation} \label{EQ8}
\begin{aligned}	
L_{clu}=\sum_{i} \sum_{t} p_{i t} \log \frac{p_{i t}}{q_{i t}},
\end{aligned}
\end{equation}	
where $p_{i t}=\frac{q_{i t}^{2} / \sum_{i} q_{i t}}{\sum_{t}\left(q_{i t}^{2} / \sum_{i} q_{i t}\right)}$ $q_{it} $ is defined as \cref{EQ6}.

With regard to network-level optimization, the reconstruction loss that plays a regularizing role is defined as
	\begin{equation} \label{Eq1}
\begin{aligned}	
L_{rec}=\frac{1}{n} \sum_{i=1}^{n}\left\|\tilde{x}_{i}-g_{\mathbf{u}}\left(f_{\mathbf{w}}\left(\tilde{x}_{i}\right)\right)\right\|_{2}^{2},
\end{aligned}
\end{equation}
where $g_{\mathbf{u}}(f_{\mathbf{w}}(\tilde x_i))$ represents the reconstructed sample by T-Encoder.

%To satisfy the preference of clustering behavior on data dimensionality, we design a loss function $L_{dim}$ to achieve dimension reduction.  

%To preserve as much of the significant structure of the features $z_{\mathbf{w}}$ as possible while satisfying the clustering preference for feature dimensionality, 

% we utilize the pairwise similarities modeling strategy and the cost function aimed at matching the joint probabilities of pairs of features in the high-dimensional and the low-dimensional spaces, used in t-SNE\cite{TSNE},  to define the dimensionality reduction loss $L_{dim}$. 
%Specifically, by converting the Euclidean distances between features into probabilities to represent similarities and minimizing a  single Kullback-Leibler divergence between a joint probability distribution, $\mathbf{\Phi}$,  in the high-dimensional space and a joint probability distribution, $\mathbf{U}$, in the low-dimensional space, we define the dimensionality reduction loss as 

To satisfy the clustering behavior preference for input dimensions and to retain as much potentially important information in the feature space $\mathcal Z_w$ as possible, we define the dimension reduction loss function $L_{dim}$ that aims to match the joint probability of each feature pair in the feature space $\mathcal Z_w$ from the T-Encoder and the clustering space $\mathcal Z_v$ of the Dim-Reduction block. $\mathcal Z_v$ is derived from space $\mathcal Z_w$ but with much lower dimensionality than $\mathcal Z_w$, hence is clustering-friendly. Specifically, by converting the Euclidean distances between features into probabilities to represent similarities and minimizing the difference between two joint probability distributions $\mathbf{\Phi}$ and $\mathbf{\Omega}$, from $\mathcal Z_w$ and $\mathcal Z_v$ respectively.
\begin{equation} 
\label{EQ11} 
\begin{aligned} L_{dim}=K L(\mathbf{\Phi} | \mathbf{\Omega})=\sum_{i} \sum_{j} \phi_{i j} \log \frac{\phi_{i j}}{\omega_{i j}}. 
\end{aligned} 
\end{equation}
The joint probabilities $\omega_{i j}$ in clustering space $\mathcal Z_v$ are defined as 
\begin{equation}
\label{EQ12}
\begin{aligned}	
\omega_{i j}=\frac{\left(1+\left\|z_{\mathbf{v}}^{i}-z_{\mathbf{v}}^{j}\right\|^{2}\right)^{-1}}{\sum_{a \neq b}\left(1+\left\|z_{\mathbf{v}}^{a}-z_{\mathbf{v}}^{b}\right\|^{2}\right)^{-1}}
\end{aligned}
\end{equation}
utilizing a Student t-distribution. In feature space $\mathcal Z_w$, however, the effects of outliers are addressed by employing symmetrized conditional probabilities. Here, $\phi_{i j}$ is calculated as $\frac{\phi_{j \mid i}+\phi_{i \mid j}}{2 n}$, with $\phi_{j \mid i}$ being defined as 
\begin{equation} \label{EQ13}
\begin{aligned}	
\phi_{j \mid i}=\frac{\exp \left(-\left\|z_{\mathbf{w}}^{i}-z_{\mathbf{w}}^{j}\right\|^{2} / 2 \sigma_{i}^{2}\right)}{\sum_{j^{\prime} \neq i} \exp \left(-\left\|z_{\mathbf{w}}^{i}-z_{\mathbf{w}}^{j^{\prime}}\right\|^{2} / 2 \sigma_{i}^{2}\right)},
\end{aligned}
\end{equation}
where $\sigma_i$ is the variance of the Gaussian that is centered on feature $z_{\mathbf{w}}^{i}$, and its selection principle is the same as t-SNE. From an algorithmically robust view, here we bind the reconstruction loss and dimension reduction loss together and unify them as structure loss.
 	\begin{equation} \label{EQ10}
\begin{aligned}	
L_{stru}=L_{rec}+\beta*L_{dim} 
\end{aligned}
\end{equation}
where $\beta>0$ are the coefficients used to balance the T-Encoder and the Dim-Reduction block respectively and is fixed at 0.001 in this work. Instead of optimizing separately, the merging of the two loss terms is intended to allow TDEC to simultaneously learn embedded features with global dependencies and to be well clustered.

 Based on the above formulations, now the overall objective function combines $L_{clu}$, $L_{dim}$ and $L_{rec}$ as 
	\begin{equation} \label{EQ7}
\begin{aligned}	
L = L_{stru} + \alpha * L_{clu}
\end{aligned}
\end{equation}
where $\alpha$ > 0 is a coefficient that controls the degree of involvement of clustering objective. Note that $\alpha$ is set to 0.1 in this work.
 
%The TDEC involves a two-stage training process comprising pre-training and fine-tuning. During pre-training, the model is trained using the net loss $L_{net}$ to extract generalized features from the input data. In the fine-tuning stage, the model is further trained by incorporating both the net loss $L_{net}$ and the clustering loss $L_{clu}$ in the global loss $L$. The objective is to utilize the clustering loss to constrain the net loss and to apply the features learned by the model to the clustering algorithm. The results obtained from the clustering algorithm are then utilized to guide the optimization of the model's features, making it more suitable for the clustering task.

We first pretrain the parameters of TDEC by setting $\alpha=0$, only using the structure loss $L_{stru}$, to extract get meaningful features of dataset $X$. Subsequently, the TDEC model is fine-tuned by minimizing the total loss expressed in \cref{EQ7}. In practice, TDEC tends to be stable after several iterations.

%We further fine-tune the TDEC model by fusing L1, L2 and L3.
%
%get meaningful target distribution.
%
% After pretraining, the cluster centers are initialized by performing k-means on embedded features of all images. Then set γ = 0.1 and
%update CAE’s weights, cluster centers and target distribution P as follows.

\begin{figure}[H]
 	%\begin{figure}[!htp]
 	\centering
 	%	\addtocounter{subfigure}{6}
 	\subfloat[Visualization of the embedded representations at 5 iterations.]
 	{\includegraphics[width=0.495\linewidth]{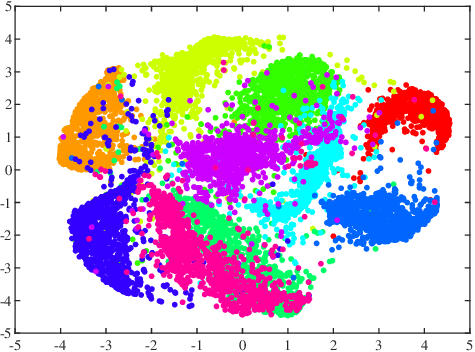}} \hspace{0mm}
 	\subfloat[Visualization of the embedded representations at 100 iterations.]
 	{\includegraphics[width=0.495\linewidth]{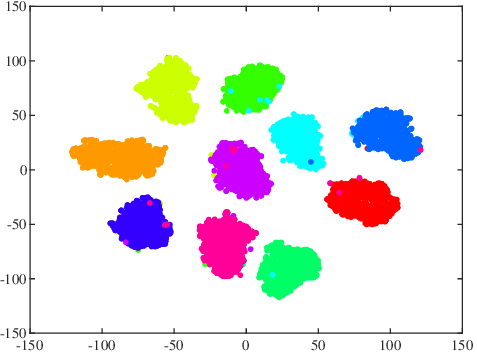}} \vspace{-3mm} 
 	\subfloat[The total loss VS training epochs]
 	{\includegraphics[width=0.7\linewidth]{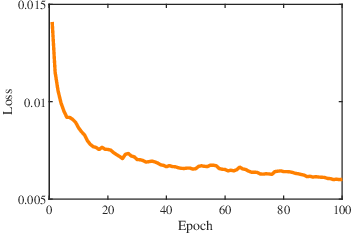}}  \vspace{-3mm}
 	\subfloat[The ACC and NMI VS training epochs]
 	{\includegraphics[width=0.7\linewidth]{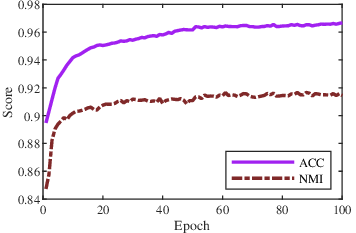}} \vspace{-3mm} 
 	\caption{The convergence process of our model on MNIST}
 	%	\caption{The heat distribution graph of density under different parameter value $k$.}
 	\label{Fig4}
 \end{figure}

We depict the convergence process of TDEC by visualizing the learned features within the different optimization stages as shown in \cref{Fig4}, where different clusters are rendered with different colors. The result in \cref{Fig4}(a),(b) shows that, at the beginning, original features are all mixed and unguided, with several clusters squeezed together. As the training process proceeds, the formed clusters become more reasonable and features scatter more distinctly. At the 100th iteration, our model reaches stability in \cref{Fig4}(c). Consequently, the clustering performance continuously increases in each round, as shown in \cref{Fig4}(d). The training flows of the proposed TDEC is presented in \cref{Algorithm1}. 
 
  	\begin{algorithm}[htp]
 	\DontPrintSemicolon
 	\SetAlgoLined
 	\KwIn {Dataset $X$, Number of clusters $K$, batch size $B$, Maximum iterations $MaxIter$, and Stopping threshold $\epsilon$}
 	\KwOut {clustering result $\mathcal Y$ }
 	Initialize weights $w$,$u$, $v$ by minimizing $L_{stru}$ in \cref{EQ10}; \\
 	\For{iter=1 to MaxIter}{
 		% {\color{red}	Feed image into T-Encoder and Dim-reduction block to learn embedded representation $Z_v$ via Eq. 1 and Eq. 2;}
 		Compute all clustering features $\{z_{\mathbf{v}}^i=h_{\mathbf{v}}(f_{\mathbf{w}}(\tilde{x}_i))\}_{i=1}^n$; \\
 		Update centers $\{c_t\}_{t=1}^K$ by density peaks via \cref{EQ4}; \\
 		Save last label assignment $\mathcal{Y'}=\mathcal{Y}$ ;   \\ 
 		Compute new label assignment $\{\mathcal{Y}_i=\mathop{argmax}\limits_{t} q_{it}\}_{i=1}^n$ via \cref{EQ6};  \\
 		% Estimate loss1, loss2 and loss3 via Eq. 1 and Eq. 2 and Eq. 3; \\
 		% Save the last label $\mathcal{Y'}$ \\
 		% Calcluate the new label $\mathcal{Y}$ \\ 
 		\If{ $\frac{sum(\mathcal{Y'}\neq\mathcal{Y})}{n}\leq \epsilon$}{  
 			\textbf{End For} 
 		}
 		Sample a mini-batch $\{x_i\}_{i=1}^{B}$ from $X$. \\
 		Update weights $w$,$u$, $v$ via \cref{EQ1}, \cref{EQ8}, \cref{EQ11}, \cref{EQ10} and \cref{EQ7} on the mini-batch.\\
 	}
 	Forward pass to obtain the final updated cluster labels; \\
 	\KwResult{$\mathcal Y$ }
 	\caption{TDEC}
 	\label{Algorithm1}
 \end{algorithm}
 
\section{Experiments}\label{Experiments}

In this section, we evaluate our proposed TDEC in several ways. To begin with, we compare it to a variety of classical and recent SOTA algorithms. Afterward, we take a closer look at the structural properties of TDEC through ablation experiments. Additionally, we test the robustness of our approach by studying how different parameters affect clustering performance.

\textbf{Evaluation Metrics.} To measure the performance of our approach, we use two standard metrics: clustering accuracy (ACC) and Normalized Mutual Information (NMI). The higher the values of these metrics, the better the performance.

%\textbf{Data Sets.} We run our experiments on image datasets MNIST\cite{MNIST}, USPS\cite{USPS}, Fashion MNIST (F-MNIST)\cite{FMNIST}, Emnist-Letters\cite{Letter}, Digits\cite{Digits}, traffic sign dataset GTSRB\cite{GTSRB}, and face dataset YTF\cite{YTF}. The experiment details and processing of the datasets are provided in our supplementary material. To improve the data complexity, for the GSTB dataset, the first five largest clusters are selected in this paper; for the YTF dataset, we select the first 80 classes from 1595 classes of images in alphabetical order.

\textbf{Data Sets.} We run our experiments on 10 image datasets, where the relevant information for these datasets is provided in \cref{Table1}. In particular, to improve the data complexity, the first five largest clusters are selected for the traffic sign dataset GRSRB; for the face dataset YTF, we select the first 80 classes from 1595 classes in alphabetical order; we utilize the ResNet50 \cite{resnet} to extract 2048-dimensional features from STL-10 datasets.

\textbf{Experimental Setup.} We construct our TDEC model with the Encoding-layer of architecture 512-512-3072-10 fully-connected network, the Decoding-layer symmetric to the Encoding-layer, and the Dim-Reduction block of architecture 10-50-50-100-2 fully-connected network. In particular, the values of $N_1$ and $N_2$ in the TDEC model are fixed at 4 and 1, respectively. We first pre-train 200 epochs by a TDEC without Clustering Head, then fit the pre-trained weight to initialize our model. The Adam optimizer\cite{adam} is used in our method. The learning rate is set as 0.01, the training epochs are set to 500, the batch size is fixed at 256, and the clusters $K$ are given by the categories of the corresponding dataset. As for the settings of hyper-parameter, $\alpha$ and $\beta$ are fixed as 0.1 and 0.001, and we further discuss the impact of different values of $k$ standing for the number of nearest neighbors on clustering in \cref{Sensitivity}.    Additional configuration details of the comparison algorithms can be found in our supplementary material.

 \begin{table}[H]
	\caption{Dataset statistics.}
	\label{Table1}
	\renewcommand\tabcolsep{1pt}
	\renewcommand\arraystretch{1}
	%	\footnotesize
	\begin{tabular}{ccccc}
		\toprule
		Dataset    & \# Samples & \# size  & \# classes   & \# type\\
		\midrule
		MNIST\tablefootnote{http://yann.lecun.com/exdb/mnist/} &    70000      &      1*28*28        &    10  & Large-Scale\\
		MNIST-test\footnotemark[1] &    10000      &      1*28*28        &    10 & --\\
		Fashion\cite{FMNIST}    &    70000      &      1*28*28        &    10 & Large-Scale\\
		USPS\cite{USPS}       &    9298       &      1*16*16        &    10 & -- \\
		Digits\tablefootnote{https://archive.ics.uci.edu/ml/machine-learning-databases/optdigits/}     &    1797       &      1*8*8          &    10 & Small-scale\\
		GTSRB\tablefootnote{http://benchmark.ini.rub.de}      &    8790       &      3*32*32        &    5 & Complex Contextual\\
		YTF\tablefootnote{http://www.cs.tau.ac.il/wolf/ytfaces/}        &    20000      &      3*64*64        &    80 & Multi-Cluster, Video\\
		Letter\cite{Letter}     &    56000     &      1*28*28        &    10 & Large-Scale \\
		STL-10\tablefootnote{https://cs.stanford.edu/ acoates/stl10/}      &    13000       &      3*96*96        &    10 & Complex Contextual\\
		Reuters-10k\tablefootnote{https://keras.io/api/datasets/reuters/}      &    10000       &      1200        &    4 & Complex Text dataset\\
		\bottomrule   
	\end{tabular}
\end{table}

\subsection{Qualitative Study}	\label{Qualitative}
%The performance comparison with classical methods and state-of-the-art DC basslines are shown in \cref{Table2}, where it is observed that TDEC yields superior performance on two different types of datasets. Especially on MNSIT, the proposed method outperforms LNSCC (the latest SOTA model) by 2.1\% and 3.6\% in terms of ACC and NMI. Whereas, on the other two datasets (MNIST-Test and FASHION), TDEC also consistently maintains the top two scores.  Additionally, \cref{Table3} reports the comparison with recent SOTA DEC-based approaches. Our method significantly prevails these state-of-the-art baselines by a large margin on four datasets. It is worth noting that the GTSRB is contextually complex, including a wide variety of traffic sign images in various climates; the Digits is intuitively significantly smaller than the volume required for convergence of common network models; whereas the YTF has a total of 80 face clusters. In particular, \cref{Fig5} visualize the top 20 scoring images closest to the cluster center in each cluster from the results of TDEC on GTSRB. \cref{Fig6} shows the clustering results for the YTF, with 15 of the 80 clusters. Moreover, TDEC surpasses the closest competitor EDESC by 4\% on YTF, 12\% on Digits, and 27\% on Letter in terms of NMI. 

\begin{figure}[H]
	\centering
	{\includegraphics[width=0.9\linewidth]{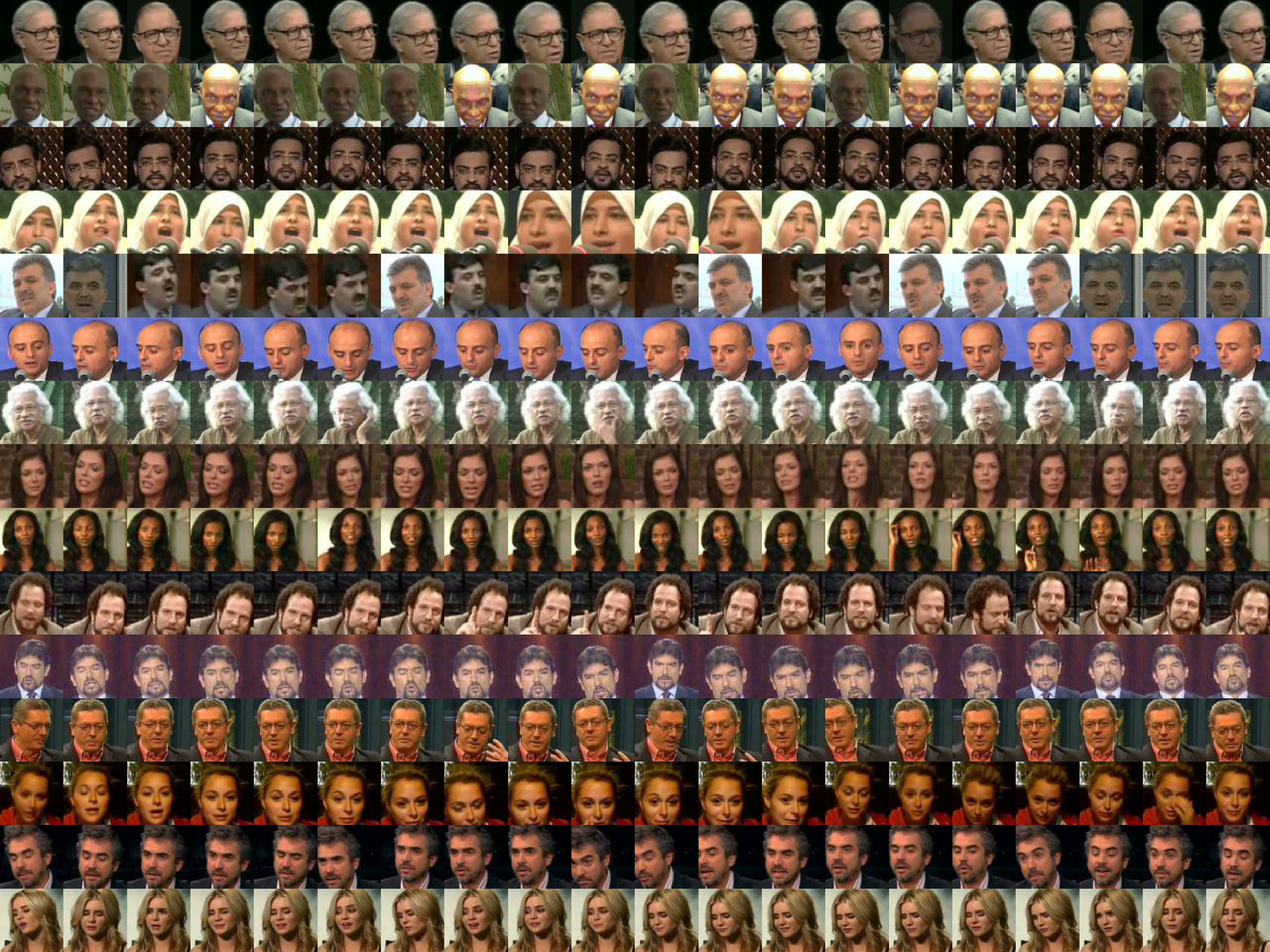}} 
	\caption{Clustering samples on the YTF dataset.Each row contains the top 20 scoring images from one cluster, based on the distance from the cluster center.}
	\label{Fig6}
\end{figure}

\begin{figure}[H]
	\centerline{
		{\includegraphics[width=1\linewidth]{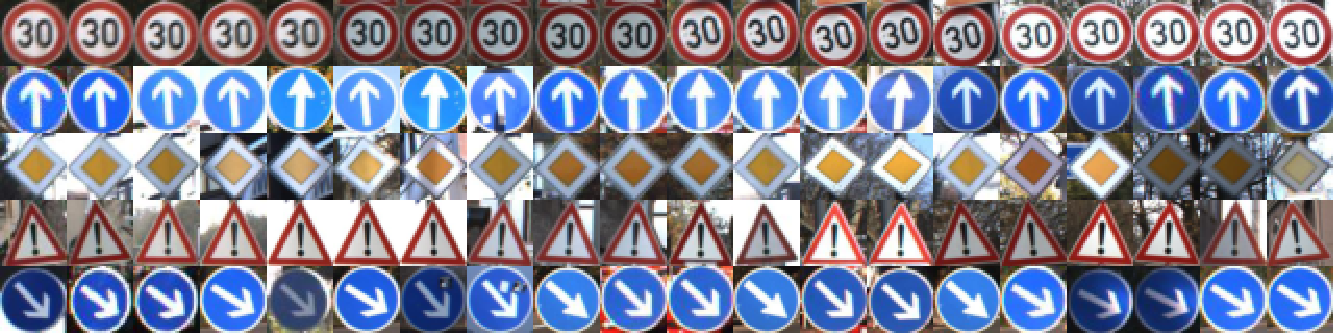}} \,}
	\caption{Clustering samples of TDEC on GTSRB. Each row contains the top 20 scoring images from one cluster, based on the distance from the cluster center.}
	\label{Fig5}
\end{figure}

The performance comparison with classical methods and state-of-the-art DC basslines are shown in \cref{Table2}, where it is observed that TDEC yields superior performance on two different types of datasets. Especially on MNSIT, the proposed method outperforms LNSCC (the latest SOTA model) by 2.1\% and 3.6\% in terms of ACC and NMI. Additionally, \cref{Table3} reports the comparison with recent SOTA DEC-based approaches. Our method significantly prevails over these state-of-the-art baselines by a large margin on four datasets. It is worth noting that the GTSRB is contextually complex, including a wide variety of traffic sign images in various climates; the Digits is intuitively significantly smaller than the volume required for convergence of network model, whereas the YTF has a total of 80 face clusters. In particular, \cref{Fig5} visualize the top 20 scoring images closest to the cluster center in each cluster from the results of TDEC on GTSRB. \cref{Fig6} shows the clustering results for the YTF, with 15 of the 80 clusters. Moreover, TDEC surpasses the closest competitor EDESC by 4\% on YTF, 12\% on Digits, and 27\% on Letter in terms of NMI. Whereas, on the two well-known challenging datasets (STL-10 and Reuters-10k), TDEC consistently maintains the top two scores.

%The STL-10 is acquired from real environments where the images have diverse backgrounds and are commonly multi-scale or multi-objective;

%the well-known complex data sets STL and REU.

Generally, the satisfactory performance of TDEC is mainly thanks to the significant discriminative property of the feature fusion in the global view and the robust assignment of the embedded distribution information in a friendly low-dimensional clustering space.

\begin{table*}[htp]
	\caption{Clustering performance compared with similar methods (state-of-the-art deep embedded approaches) in terms of ACC(\%) and NMI(\%, in parenthesis). The bolded font represents the best and second results.}
	\label{Table3}
	\renewcommand\tabcolsep{8pt}
	\renewcommand\arraystretch{1}
	\centerline{
		\begin{tabular}{cccccccc}
			\toprule
			Method  & MNIST               & GTSRB               & Digits              & YTF                 & Letter & STL-10 & Reuters-10k          \\
			\midrule
			DEC\cite{DEC}     & 84.9(81.6)          & 65.7(55.1)          & 72.3(69.7)          & 71.5(93.8)          & 53.1(51.5)     & 35.9(27.6) & 73.7(49.7)     \\
			IDEC\cite{IDEC}    & 88.1(86.7)          & 69.9(60.8)          & 76.4(71.6)          & 70.0(93.2)          & 54.1(52.0)     & 37.8(32.4) & 75.6(49.8)     \\
			DipDECK\cite{DipDECK} & 96.1(90.3)          & 61.9(49.5)          & \textbf{88.3(83.2)}          & 79.7(92.6)          & 52.5(\textbf{59.2})   & -(-) & -(-)       \\
			DeepDPM\cite{DeepDPM} & \textbf{98.0(94.0)}          & 79.9(66.9)          & 85.4(77.9)          & 82.1(93.0)          & \textbf{61.2}(58.3)    & \textbf{85.0(79.0)} & \textbf{83.0}(61.0)      \\
			IDCEC\cite{IDCEC}   & 94.8(90.6)          & -(-)                & -(-)                & -(-)                & -(-)          & -(-) & -(-)      \\
			EDESC\cite{EDESC}   & 91.3(86.2)          & \textbf{85.0(67.3)} & 84.0(79.5)          & \textbf{85.5(93.9)}          & 49.9(43.2)     & 74.5(68.7)    & \textbf{82.5(61.1)} \\
			Ours    & \textbf{98.5(95.7)} & \textbf{82.4(71.5)}          & \textbf{95.5(91.0)} & \textbf{95.0(98.0)} & \textbf{68.0(70.3)} & \textbf{85.5(77.5)} & 81.7(\textbf{62.3})\\
			\bottomrule 
		\end{tabular}
	}
\end{table*}

\begin{table}
	\caption{Clustering performance compared with other DC methods (baseline and state-of-the-art approaches) in terms of ACC(\%) and NMI(\%, in parenthesis). The bolded font represents the best and second results.}
	\label{Table2}
	\renewcommand\tabcolsep{2pt}
	\renewcommand\arraystretch{1}
	\centerline{
		\begin{tabular}{ccccc}
			\toprule
			Method      & MNIST      & MNIST-Test & FASHION    & USPS       \\
			\midrule
			k-means\cite{k-means}     & 53.2(50.0) & 54.6(50.1) & 47.4(51.2) & 66.8(62.7) \\
			SC\cite{SC}          & 65.6(73.1) & 66.0(70.4) & 50.8(57.5) & 64.9(79.4) \\
			AC\cite{AC}          & 62.1(68.2) & 69.5(71.1) & 50.0(56.4) & 68.3(72.5) \\
			GMM\cite{GMM}         & 43.3(36.6) & 54(49.3)   & 55.6(55.7) & 55.1(53.0) \\
			JULE\cite{JULE}        & 96.4(91.3) & 96.1(91.5) & 56.3(60.8) & 95.0(91.3) \\
			DEPICT\cite{DEPICT}      & 96.5(91.7) & 96.3(91.5) & 39.2(39.2) & 89.9(90.6) \\
			Deepcluster\cite{Deepcluster} & 79.7(66.1) & 85.4(71.3) & 54.2(51.0) & 56.2(54.0) \\
			CCc\cite{CC}          & 88.6(82.0) & -(-)       & -(-)       & 80.6(74.8) \\
			ASPC-DA\cite{ASPC-DA}     & 85.9(84.2) & 78.2(73.4) & 60.0(63.3) & 75.3(76.6) \\
			VaGAN-SMM\cite{VaGAN-SMM}   & 95.7(90.1) & -(-)       & 66.9(67.1) & -(-)       \\
			DSEC\cite{DSEC}        & \textbf{98.3(95.2)} & -(-)       & -(-)       & -(-)       \\
			LNSCC\cite{LNSCC}       & 96.4(92.1) & \textbf{98.2(96.9)} & \textbf{66.5(70.3)} & \textbf{97.0(93.7)} \\
			HC-MGAN\cite{HC-MGAN}     & 94.3(90.5) & -(-)       & \textbf{72.1}(69.1) & -(-)       \\
			Ours        & \textbf{98.5(95.7)} & \textbf{97.5(93.5)} & 64.5\textbf{(69.3)} & \textbf{97.6(93.5)} \\
			\bottomrule 
		\end{tabular}
	}
\end{table}

\subsection{Ablation Study}	\label{Ablation}
\begin{figure}[h]
	%\begin{figure*}[!htp]
	\centering
	%	\addtocounter{subfigure}{6}
	\subfloat[]
	{\includegraphics[width=0.45\linewidth]{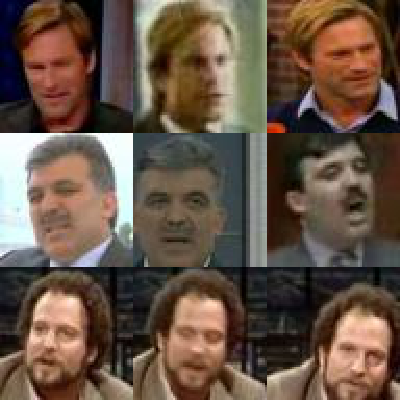}}\,
	\subfloat[]
	{\includegraphics[width=0.45\linewidth]{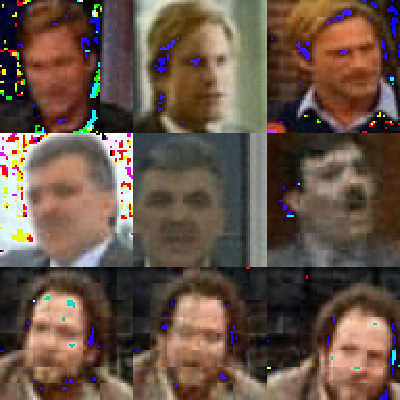}} \\
	\caption{Output results of T-Encoder and T-Decoder on YTF dataset. (a): the raw images of three individuals in each of the 3 poses; (b): the corresponding reconstructed images.}
	%	\caption{The heat distribution graph of density under different parameter value $k$.}
	\label{Fig7}
\end{figure}

We conduct an ablation study to further understand the importance of the Transformer blocks (ST), Clustering Head (CH), and Dim-Reduction block (DR) in the proposed method on the clustering performance. Specifically, we construct three degradation models by removing the corresponding modules (blocks). \cref{Table3} summarizes the results of the ablation study, from which we can draw some conclusions. First, the Transformer block is crucial to capture discriminative representations that are beneficial to fit the clustering context, as a significant performance improvement can be observed after adding the Transformer block. Particularly, \cref{Fig7} illustrates the reconstructed image of TDEC on YTF using Transformer block, implying the importance of Transformer block for clustering complex image. The general AE structure is tough to achieve such performance even with the additional convolution operation. Second, the clustering Head incorporating the distribution information could help the model to obtain high-confidence assignment results for the training network. Third, the DR module is important to meet the preference of the clustering behavior on dimensionality and simultaneously maintain the joint training of the embedded features. Interestingly, \cref{Table4} demonstrates that the clustering performance is progressively improving, which suggests the joint effects of the three modules to some extent.

\begin{table}[htp]
	\caption{Ablation study results of the proposed method and its
		degradation models.}
	\label{Table4}
	\renewcommand\tabcolsep{2pt}
	\renewcommand\arraystretch{1}
	%  \footnotesize
	\begin{tabular}{cccccccc}
		\toprule
		  ST & CH & DR & MNIST   & USPS   & Digits  & Letter   & GTSRB \\
		\midrule
	 $\times$ & $\times$ & $\times$ & \begin{tabular}[c]{@{}c@{}}ACC:0.84\\ NMI:0.81\end{tabular} & \begin{tabular}[c]{@{}c@{}}ACC:0.74\\ NMI:0.75\end{tabular} & \begin{tabular}[c]{@{}c@{}}ACC:0.52\\ NMI:0.52\end{tabular} 
		& \begin{tabular}[c]{@{}c@{}}ACC:0.53\\ NMI:0.51\end{tabular}
		& \begin{tabular}[c]{@{}c@{}}ACC:0.66\\ NMI:0.55\end{tabular}\\
		\midrule
		 \checkmark & $\times$ & $\times$ & \begin{tabular}[c]{@{}c@{}}ACC:0.92\\ NMI:0.84\end{tabular} & \begin{tabular}[c]{@{}c@{}}ACC:0.82\\ NMI:0.80\end{tabular} & \begin{tabular}[c]{@{}c@{}}ACC:0.83\\ NMI:0.83\end{tabular} 
		& \begin{tabular}[c]{@{}c@{}}ACC:0.60\\ NMI:0.59\end{tabular}
		& \begin{tabular}[c]{@{}c@{}}ACC:0.72\\ NMI:0.63\end{tabular}\\
		\midrule
		 \checkmark & \checkmark & $\times$ & \begin{tabular}[c]{@{}c@{}}ACC:0.93\\ NMI:0.86\end{tabular} & \begin{tabular}[c]{@{}c@{}}ACC:0.90\\ NMI:0.83\end{tabular} & \begin{tabular}[c]{@{}c@{}}ACC:0.87\\ NMI:0.80\end{tabular}
		& \begin{tabular}[c]{@{}c@{}}ACC:0.65\\ NMI:0.67\end{tabular}
		& \begin{tabular}[c]{@{}c@{}}ACC:0.74\\ NMI:0.68\end{tabular}\\
		\midrule
	 \checkmark & \checkmark & \checkmark & \begin{tabular}[c]{@{}c@{}}ACC:\textbf{0.97}\\ NMI:\textbf{0.93}\end{tabular} & \begin{tabular}[c]{@{}c@{}}ACC:\textbf{0.97}\\ NMI:\textbf{0.93}\end{tabular} & \begin{tabular}[c]{@{}c@{}}ACC:\textbf{0.95}\\ NMI:\textbf{0.91}\end{tabular} 
		& \begin{tabular}[c]{@{}c@{}}ACC:\textbf{0.68}\\ NMI:\textbf{0.70}\end{tabular}
		& \begin{tabular}[c]{@{}c@{}}ACC:\textbf{0.82}\\ NMI:\textbf{0.71}\end{tabular}\\
		\bottomrule
	\end{tabular}
\end{table}

\subsection{Parameter Sensitivity}\label{Sensitivity}
\begin{figure}[htp]
	\centerline{
		{\includegraphics[width=0.95\linewidth]{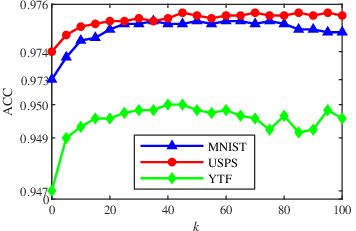}} \,}
	\caption{Robustness of TDEC to different $k$, which can be derived that 1) $k$ has good stability and 2) the neighborhood information is more favorable than the traditional distance information for clustering.}
	\label{Fig8}
\end{figure}

In \cref{Fig8}, we investigate the impact of hyper-parameter $k$ on the clustering performance of YTF, MNIST, and USPS datasets. The philosophy behind $k$ is to dynamically incorporate neighborhood information in the assignment process, i.e., to estimate the exact distance between the target object and each cluster using its neighbors instead of itself. For the trained model, we set the value range of $k$ to [0,100], where it is worth mentioning that $k=0$ means that there are no neighbors to assist the assignment process, i.e., it degenerates to the current widely used one-to-one pairwise-distance. \cref{Fig8} supports the following observations. First, leveraging neighborhood information in the assignment process does boost the clustering performance, especially on YTF, which illustrates that the neighborhood information is beneficial for handling challenging scenarios such as multiple-cluster. Moreover, although hyper-parameters are added, the optimal domain of $k$ is wide and generally distributed between 30 and 60. In our proposed method, $k$ is fixed to 50 across all datasets, i.e., no fine-tuning is provided per dataset.

\section{Conclusion}\label{Conclusion}
 In this paper, we proposed a new deep image clustering, TDEC, which is applicable to challenging scenarios such as multiple-cluster, small-scale, large-scale, and complex backgrounds. The proposed T-Encoder consists of a stack of Transformer blocks that is capable of capturing the core information of the complex image. Moreover, the Dim-reduction block bridges the gap between the learning and clustering process in terms of feature dimensionality. Finally, the low-dimensional clustering space and the objective distribution information jointly enhance the final clustering performance. The qualitative study, ablation experiment, and robustness test demonstrate the effectiveness of TDEC over state-of-the-art competitors.

\bibliographystyle{ACM-Reference-Format}
\bibliography{main}

\clearpage
\onecolumn
\appendix
\section*{Appendix}

In this appendix, we first present the details of the datasets used in our experiments and the details of the processing of each dataset, along with the data enhancement approach (Section 1). Then, we describe in detail the data sources for the comparison algorithm (Section 2). In addition, the computational procedure for the metrics used during the experiments is given in detail (Section 3).

\section{Datasets}
A summary of the datasets used in our experiments is given in \cref{app:tab-datasets}.

\begin{table}[H]
	\caption{Dataset statistics.}
	\label{app:tab-datasets}
	\renewcommand\tabcolsep{3pt}
	\renewcommand\arraystretch{1}
	\begin{tabular}{ccccc}
		\toprule
		Dataset & \# Samples & \# image size & \# classes & \# type\\
		\midrule
		MNIST\cite{MNIST} & 70000 & 1*28*28 & 10 & Large-Scale\\
		MNIST-test\cite{MNIST} & 10000 & 1*28*28 & 10 & Unbalanced\\
		Fashion\cite{FMNIST} & 70000 & 1*28*28 & 10 & Large-Scale\\
		USPS\cite{USPS} & 9298 & 1*16*16 & 10 & Unbalanced\\
		Digits\cite{Digits} & 1797 & 1*8*8 & 10 & Small-scale\\
		GTSRB\cite{GTSRB} & 8790 & 3*32*32 & 5 & Complex contextual\\
		YTF\cite{YTF} & 20000 & 3*64*64 & 80 & Multi-Cluster\\
		Letter\cite{Letter} & 56000 & 1*28*28 & 10 & Large-Scale\\
		\bottomrule
	\end{tabular}
\end{table}

\noindent\textbf{MNIST} \cite{MNIST}: Greyscale image data set consisting of 7000 handwritten digits (0 to 9) with a size of 28 × 28 pixels. The number of images in the training set is 60,000, and the number of images in the test set is 10,000.\\
\textbf{Fashion-MNIST} (F-MNIST) \cite{FMNIST}: Greyscale image data set consisting of 70000 articles from the Zalando online store. Each sample belongs to one of 10 product groups and has a size of 28 × 28 pixels.\\
\textbf{USPS} \cite{USPS}: Greyscale image data set consisting of 9298 handwritten digits (0 to 9) with a size of 16 × 16 pixels.\\
\textbf{Digits} \cite{Digits}: Greyscale image data set consisting of 1797 handwritten digits (0 to 9) with a size of 8 × 8 pixels.\\
\textbf{GTSRB} \cite{GTSRB}: Color image data set consisting of 8790 traffic signal boards (a total of 5 categories) with a size of 3 × 32 × 32 pixels. In order to ensure balance among the number of data sets, we chose the five categories with the highest number of data sets from all available options.\\
\textbf{YTF} \cite{YTF}: Color image data set consisting of 20000 face pictures (a total of 80 people) with a size of 3 × 64 × 64 pixels. In order to maintain balance in the number of sets, we selected 80 people in alphabetical order from each category. To ensure the proper size, the selected images were then cropped.\\
\textbf{Letter} \cite{Letter}: Greyscale image data set consisting of 56000 handwritten letters (A to J) with a size of 28 × 28 pixels.

\begin{figure}[H]
\centering
\subfloat[MNIST]{\includegraphics[width=3cm]{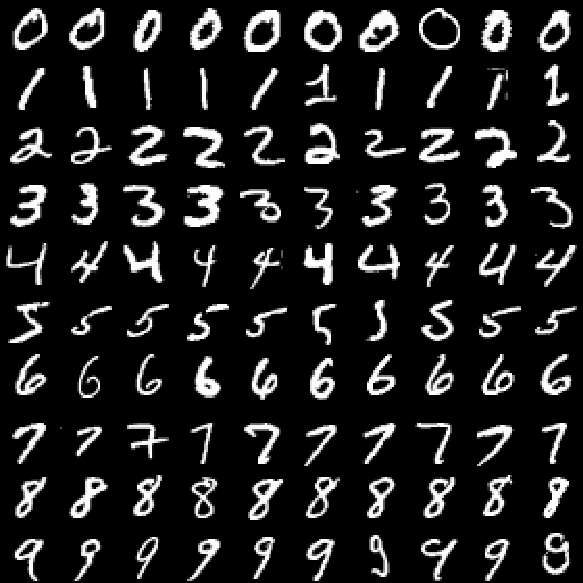}}
\quad
\subfloat[FMNIST]{\includegraphics[width=3cm]{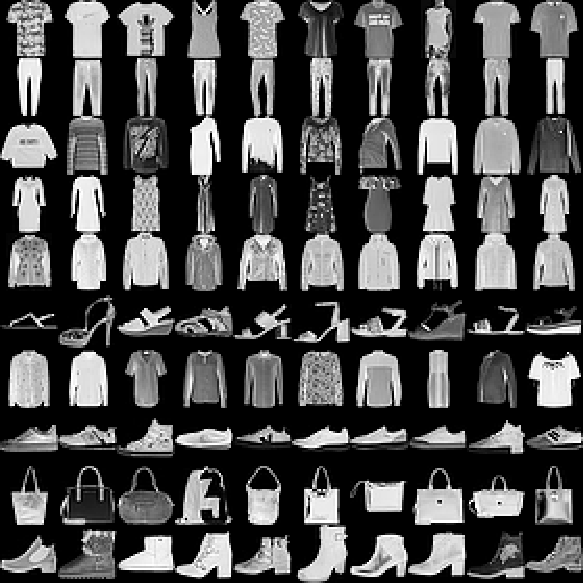}}
\quad
\subfloat[USPS]{\includegraphics[width=3cm]{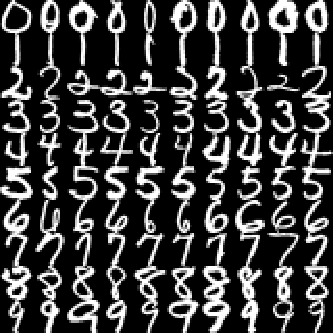}}
\quad
\subfloat[Letter]{\includegraphics[width=3cm]{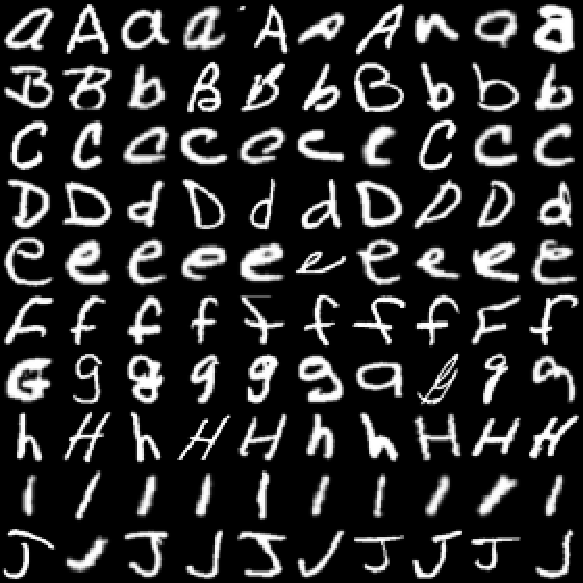}}
\quad
\subfloat[Digits]{\includegraphics[width=3cm]{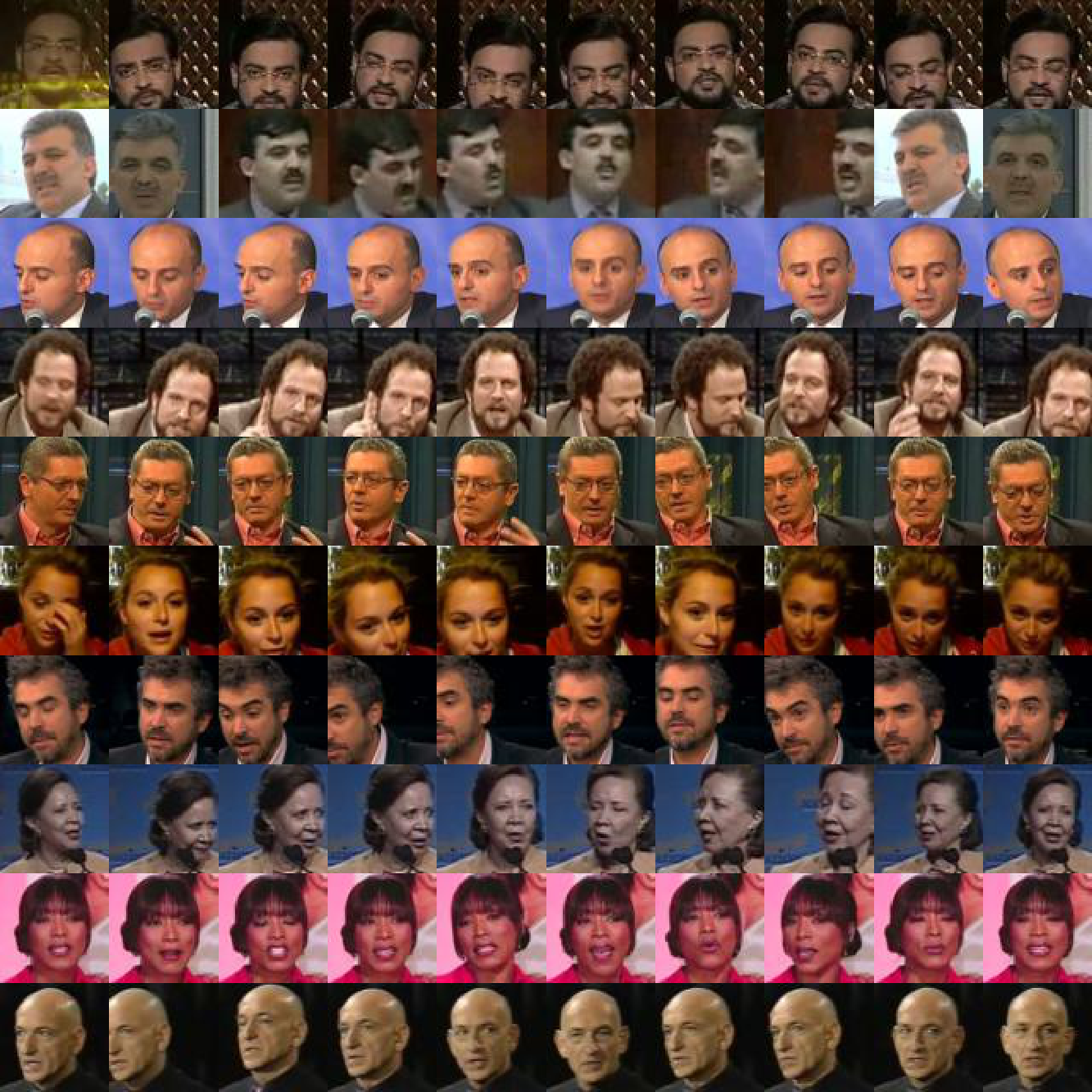}}
\quad
% \subfloat[GTSRB]{\includegraphics[width=3cm]{dataset_example_GTSRB.eps}}
\caption{Datasets visualization.}
\label{app:fig-datasets}
\end{figure}

In order to enhance the image data used during model training, we employed two methods: randomly rotating the images up to 10 degrees and shifting them by a maximum of $\frac{H}{10}$ pixels in any direction, where $H$ is the height of the input image. In addition, as the patch size for sinusoidal position encodings must be even, we performed a resize operation on a portion of the dataset, primarily converting images of size 28 × 28 to 32 × 32 using standard library functions. Examples of all datasets are shown in \cref{app:fig-datasets}.

\section{Comparison Algorithms}
We compared our proposed approach to a range of both classical and state-of-the-art clustering algorithms, including SC\cite{SC}, AC\cite{AC}, GMM\cite{GMM}, JULE\cite{JULE}, DEPICT\cite{DEPICT}, Deepcluster\cite{Deepcluster}, CC\cite{CC}, VaGAN-SMM\cite{VaGAN-SMM}, DSEC\cite{DSEC}, LNSCC\cite{LNSCC}, and HC-MGAN\cite{HC-MGAN}. The results for these algorithms were taken from the original or related papers. For the k-means algorithm\cite{k-means}, we conducted our own implementation using the scikit-learn library with a default configuration of n\_init=20 and the correct number of clusters. The results for the ASPC-DA\cite{ASPC-DA} algorithm were also obtained through our own implementation, using an AE structure of [500, 500, 1000, 10, 1000, 500, 500], a batch size of 256, 500 rounds of pre-training, and the ADAM optimizer for both pre-training and formal training. The k-means algorithm used in this experiment was also set to its default setting.

The DEC\cite{DEC} algorithm used source \href{https://github.com/XifengGuo/DEC-keras}{code-DEC} as its foundation, and the experimental setup included an AE structure of [500, 500, 2000, 10, 2000, 500, 500], a batch size of 256, and 500 pre-training rounds. The ADAM optimizer and a learning rate of 0.01 were used throughout pre-training and formal training. The IDEC\cite{IDEC} algorithm, which used source \href{https://github.com/XifengGuo/IDEC}{code-IDEC}, had the same experimental setup as DEC. For the DipDECK\cite{DipDECK} algorithm, which was based on source \href{https://dmm.dbs.ifi.lmu.de/cms/downloads/index.html}{code-DipDECK}, the initial number of clusters was set to the correct number, and all other parameters used the default settings in the source code. Similarly, the DeepDPM\cite{DeepDPM} algorithm that used source \href{https://github.com/BGU-CS-VIL/DeepDPM}{code-DeepDPM} had a setup similar to the previous DipDECK, with the exception of the initial number of clusters being set to the correct number and all other parameters using the default settings from the source code. The results for the IDCEC\cite{IDCEC} algorithm were taken from the original article. Lastly, the EDESC\cite{EDESC} algorithm, which was based on source \href{https://github.com/JinyuCai95/EDESC-pytorch}{code-EDESC}, utilized the default configuration in the source code. A brief introduction of the comparison algorithms is shown in \cref{app:tab-algorithms}.

\begin{table}[H]
	\caption{Comparison algorithms statistics.}
	\label{app:tab-algorithms}
	\renewcommand\tabcolsep{3pt}
	\renewcommand\arraystretch{1}
	\begin{tabular}{cccc}
		\toprule
		Name & \# source & \# data source & \# code source\\
		\midrule
		JULE\cite{JULE} & 2016/CVPR & original or related papers & -\\
		DEPICT\cite{DEPICT} & 2017/ICCV & original or related papers & -\\
		DeepCluster\cite{Deepcluster} & 2018/ECCV & original or related papers & -\\
		CC\cite{CC} & 2021/AAAI & original or related papers & -\\
		ASPC-DA\cite{ASPC-DA} & 2020/TKDE & our implementation & https://github.com/XifengGuo/ASPC-DA\\
		VaGAN-SMM\cite{VaGAN-SMM} & 2022/TNNLS & original or related papers & -\\
		DSEC\cite{DSEC} & 2020/TPAMI & original or related papers & -\\
		LNSCC\cite{LNSCC} & 2022/IJCAI & original or related papers & -\\
		HC-MGAN\cite{HC-MGAN} & 2022/AAAI & original or related papers & -\\
		DEC\cite{DEC} & 2016/ICML & our implementation & https://github.com/XifengGuo/DEC-keras\\
		IDEC\cite{IDEC} & 2017/IJCAI & our implementation & https://github.com/XifengGuo/IDEC\\
		DipDECK\cite{DipDECK} & 2021/SIGKDD & our implementation & https://dmm.dbs.ifi.lmu.de/cms/downloads/index.html\\
		DeepDPM\cite{DeepDPM} & 2022/CVPR & our implementation & https://github.com/BGU-CS-VIL/DeepDPM\\
		IDCEC\cite{IDCEC} & 2022/Pattern Recognition & original or related papers & -\\
		EDESC\cite{EDESC} & 2022/CVPR & our implementation & https://github.com/JinyuCai95/EDESC-pytorch\\
		\bottomrule
	\end{tabular}
\end{table}

\section{Future Work}
\label{subsec:openworld_future}

\noindent \textbf{Open-World Discovery.} 
The ability to capture global dependencies and robustly assign clusters in complex visual data is a transformative cornerstone for transitioning from static clustering to the dynamic Open-World paradigm. A major bottleneck in Generalized Category Discovery \cite{zheng2024textual} and On-the-fly Category Discovery \cite{zheng2024prototypical} is effectively representing and grouping complex, unseen visual semantics. By synergizing curriculum-driven density core assignments \cite{zheng2024deep}, and multi-modality co-teaching \cite{zheng2024textual}, future architectures can achieve a more holistic and autonomous understanding of evolving data streams. Beyond discovery, this global-aware representation capability extends naturally to data-efficient learning and generative modeling. By dynamically capturing the underlying semantic modes of complex distributions in a low-dimensional space, the proposed logic can provide critical structural guidance for adaptive dataset quantization~\cite{LiZDXQ25} and diverse dataset distillation frameworks \cite{libeyond,li2026fixed,li2026efficient}, ensuring that synthetic or compressed proxies preserve the full topological and categorical diversity of the original data. Similarly, recognizing latent cluster densities can enhance data-free knowledge distillation by anchoring diverse diffusion-based augmentations \cite{LiZ0XLQ24} to precise semantic centers. Finally, moving towards advanced 3D generative vision, these density-aware structural priors hold significant potential to refine probability density flow matching, enabling more stable and mode-preserving geodesic estimations for novel view synthesis \cite{wang2026geodesicnvs}.

\noindent \textbf{Cross-Domain Potential and Future Trajectories.}
TDEC effectively addresses the structural bottleneck of classical deep clustering by bridging discriminative feature representation with clustering-friendly latent spaces, thereby serving as a robust initialization engine for diverse downstream pipelines. For medical imaging, it facilitates pathological region partitioning by leveraging global contextual information to group distinct tissue layers or lesion types in complex OCT images, which is essential for guided denoising and segmentation \cite{li2026cross,li2025retidiff,li2024efficient,li2026garnet}. In 3D vision, TDEC enhances global solvers and vanishing point estimation \cite{zhao2026advances,zhao2024balf,zhao2023benchmark,liao2025convex} by embedding robust spatial dependencies in unstructured environments, and streamlines instance-level point cloud grouping \cite{qu2024conditional,qu2025end,qu2025robust}. For robotics, its density-based clustering head automates the identification of latent physical modes (e.g., slip, rotation, or stable contact) within noisy tactile or sensor data streams, thereby refining stability analysis and cooperative control \cite{yan2025pandas,ruan2024q,yan2023stability,zhang2025ccma,xu2025sedm,tian2025measuring}. In embodied AI and video-language modeling, TDEC aids in temporal event partitioning by clustering complex video sequences into coherent semantic events \cite{li2025lion}, and assists in defining the optimal granularity for action primitives in VLA models \cite{li2025cogvla}, which provides the structured semantic groundwork necessary for enabling deliberate System-2 reasoning capabilities \cite{song2025hume}. Furthermore, it optimizes knowledge distillation by providing highly discriminative latent representations for location-aware semantic masking \cite{lan2025acam,lan2026clockdistill,lan2024gradient}, and provides aligned latent spaces that can effectively bridge semi-supervised domain translation via diffusion models \cite{wang2025ladb}. Finally, TDEC provides a scalable way to discover entities in open-vocabulary marine or remote sensing segmentation \cite{li2025exploring,li2025stitchfusion,li2025maris,li2025exploring2} and identifies complex structural groups in socio-meteorological analysis \cite{liu2026health,shen2025aienhanced,shen2026mftformer}.

\section{Clustering Metrics}
To evaluate the clustering performance, we adopt two standard evaluation metrics: Accuracy (ACC) and Normalized Mutual Information (NMI).

The best mapping between cluster assignments and true labels is computed using the Hungarian algorithm to measure accuracy \cite{Hungarian}. For completeness, we define ACC by:
\begin{equation}
\label{app:eq-acc}
\begin{aligned}
ACC=\max _{m} \frac{\sum_{i=1}^{n} \mathbf{1}\left\{l_{i}=m\left(c_{i}\right)\right\}}{n},
\end{aligned}
\end{equation}
where $l_i$ and $c_i$ are the ground-truth label and predicted cluster label of data point $x_i$, respectively.

NMI calculates the normalized measure of similarity between two labels of the same data:
\begin{equation}
\label{app:eq-nmi}
\begin{aligned}
NMI=\frac{I(l ; c)}{\max \{H(l), H(c)\}},
\end{aligned}
\end{equation}
where $I(l, c)$ denotes the mutual information between true label $l$ and predicted cluster $c$, and $H$ represents their entropy. Results of NMI do not change by permutations of clusters (classes), and they are normalized to [0, 1] with 0 implying no correlation and 1 exhibiting perfect correlation.

\end{document}